\documentclass[runningheads]{llncs}
\usepackage[T1]{fontenc}
\usepackage{graphicx}
\usepackage{marvosym}
\usepackage{utfsym}
\usepackage{pifont}
\usepackage{amssymb}
\usepackage{amsmath}
\usepackage{multirow}
\usepackage{xcolor}
\usepackage{booktabs}
\makeatletter
\newcommand{\printfnsymbol}[1]{%
  \textsuperscript{\@fnsymbol{#1}}%
}
\makeatother
\begin{document}
\title{TRUST: Efficient Abdominal \underline{T}rauma \underline{R}ecognition via Image-to-\underline{U}ltra\underline{s}ound-Video \underline{T}ransfer learning}
\titlerunning{Image-to-Ultrasound-Video Transfer Learning}
%
%
\setcounter{footnote}{1}
\author{Enguang Wang\inst{1,3,4}\thanks{Equal Contribution.} \and
Hao Zhou\inst{2}\printfnsymbol{1} \and Shuo Gao\inst{1,3,4} \and Tuo Liu\inst{1,3,4} \and Guangquan Zhou\inst{1,3,4}\textsuperscript{(\Letter)}}
\authorrunning{Enguang Wang et al.}
%
\institute{School of Biological Science and Medical Engineering, Southeast University, Nanjing, China \\ \email{guangquan.zhou@seu.edu.cn} \and
Nanjing Medical University First Affiliated Hospital, Nanjing, China \and Jiangsu Key Laboratory of Biomaterials and Devices, Southeast University, Nanjing, China \and State Key Laboratory of Digital Medical Engineering, Southeast University, Nanjing, China}
\maketitle 
\begin{abstract}
Abdominal ultrasound is indispensable for rapid, noninvasive trauma triage. However, interpreting the subtle dynamic cues embedded in continuous scanning is time-intensive and operator-dependent. Parameter-Efficient Image-to-Video Transfer Learning (PEIVTL), which efficiently adapts pre-trained image models to the video domain, notably through visual–textual alignment, offers a promising paradigm for ultrasound video analysis. Nevertheless, substantial spatiotemporal and semantic variations arising from physician-dependent scanning practices continue to limit the effectiveness and generalizability of this framework. We propose TRUST, a scan-aware PEIVTL framework that explicitly models fine-grained spatiotemporal variations to enable reliable ultrasound video understanding. First, we introduce a Cross-Frequency Collaborative Adapter (CFCA) that establishes mutual constraints between low- and high-frequency components, enhancing discriminative spatial feature extraction under heavy speckle corruption. Second, we design a Multi-Granularity Motion-Aware (MGMA) module that integrates local temporal convolutions with motion-prior-guided global self-attention, jointly capturing stable intra-view patterns and abrupt inter-view transitions to characterize complex scanning dynamics. Third, a Visual Query Semantic Aggregation (VQSA) module dynamically generates text prototypes conditioned on visual features, enabling adaptive visual–textual alignment robust to intra-class variability under diverse scanning conditions. Experiments on in-house ultrasound trauma datasets demonstrate that TRUST outperforms state-of-the-art methods by 9.63\% with superior computational efficiency.

\end{abstract}
\section{Introduction}
In emergency trauma care, abdominal ultrasound has emerged as the preferred imaging modality for rapid triage due to its non-invasive nature, portability, and real-time capabilities~\cite{van2011abdominal}. Nonetheless, accurate trauma assessment requires clinicians to interpret subtle, often transient, dynamic cues across continuous, multi-view scanning sequences, which is highly time-intensive and operator-dependent~\cite{xu2024sammpa,zhou2025medsam}. These challenges underscore the need for automated, video-level analysis methods capable of modeling spatiotemporal information in ultrasound data. Although deep learning has achieved significant advances in medical video analysis~\cite{liu2025lovit,yang2024surgformer}, its success critically depends on large-scale, temporally annotated datasets, which remain notably scarce in the ultrasound domain~\cite{yang2025surgpetl}.  As an alternative, Parameter-Efficient Image-to-Video Transfer Learning (PEIVTL)~\cite{wang2024m2clip,wasim2023vita}, which adapts pre-trained larger models to the video domain by leveraging visual–textual alignment in vision-language frameworks such as CLIP~\cite{radford2021CLIP}, offers a promising paradigm for ultrasound video analysis.
\begin{figure}[t]
\includegraphics[width=\textwidth]{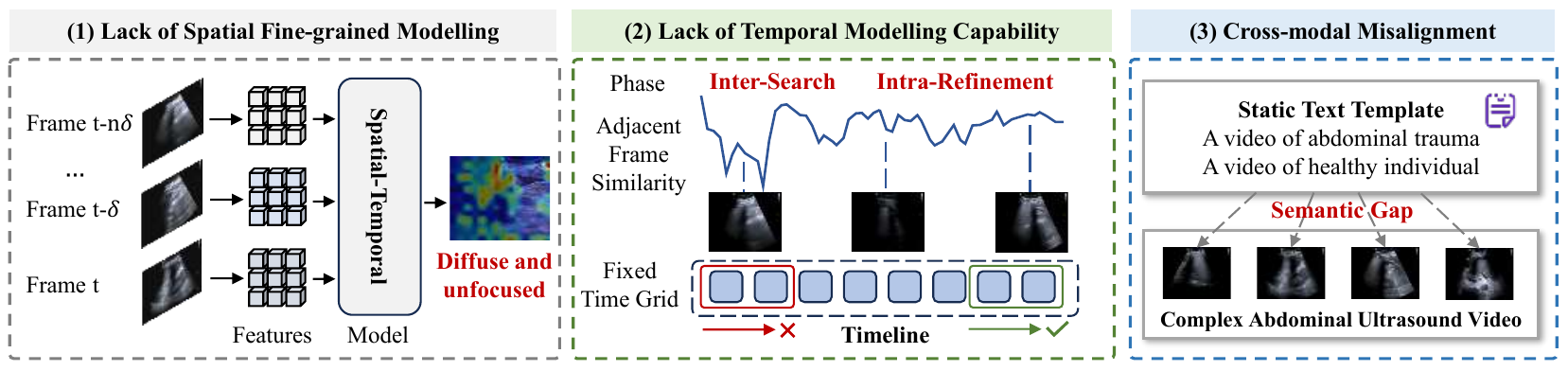}
\caption{Existing methods lack fine-grained spatial extraction, temporal dynamic modeling, and cross-modal matching capabilities in abdominal ultrasound applications.} \label{fig1}
\end{figure}

Despite the demonstrated effectiveness of PEIVTL~\cite{yang2025surgpetl,wang2024tr-adapter,pei2025d2st}, its direct application to abdominal ultrasound video analysis remains non-trivial due to the pronounced spatiotemporal and semantic variability introduced by operator-dependent scanning. First, speckle noise and imaging artifacts in ultrasound~\cite{deng2024memsam,zhang2024domesticating} routinely obscure the subtle pathological cues that are essential for accurate diagnosis, demanding more discriminative and fine-grained spatial feature modeling. Moreover, the complex inter-scan variability arising from both rapid probe repositioning and gradual anatomical changes presents substantial challenges for characterizing the dual temporal dynamics intrinsic to ultrasound examinations. Most existing PEIVTL approaches rely on single-granularity temporal modeling, typically implemented via fixed-window local convolutions~\cite{ni2022expanding,park2023dualpath,zhu2020actionsurvey1,kong2022actionsurvey2}, which are insufficient to simultaneously capture short-range motion continuity and longer-range cross-view transitions. Finally, considerable intra-class variation induced by heterogeneous scanning angles, probe pressures, and patient-specific factors limits the adaptability of static text templates~\cite{wasim2023vita,wang2025vlpa,wang2024m2clip} used for visual-textual alignment. Such rigidity can lead to suboptimal cross-modal feature alignment, thereby compromising classification robustness and generalization performance.

In this work, we propose TRUST, a scan-aware PEIVTL framework that progressively refines ultrasound video representations along three complementary axes—spatial detail, temporal dynamics, and cross-modal semantics—within a CLIP-based pipeline. Specifically, to enhance fine-grained spatial discrimination, we introduce the Cross–Frequency Collaborative Adapter (CFCA), which leverages frequency-domain interaction to model the spectral-informed complementarity between low- and high-frequency components. Building on these enriched spatial features, the Multi-Granularity Motion-Aware (MGMA) module aggregates temporal information through a dual-branch architecture: a local branch employing temporal convolutions for short-range intra-view consistency, and a global branch incorporating motion-prior position encoding to capture long-range inter-view scanning logic. The resulting visual representation then drives the Visual Query Semantic Aggregation (VQSA) module, which uses it as a query to dynamically synthesize instance-specific text prototypes from a diverse description repository via cross-modal attention, enabling adaptive visual–textual alignment robust to intra-class variability. The contributions of this paper are threefold: 1) To our best knowledge, this is the first work to adapt PEIVTL for abdominal ultrasound video analysis; 2) We propose TRUST, a unified framework that progressively integrates fine-grained spatial modeling, motion-aware temporal encoding, and visual-query-driven semantic alignment; 3) On an abdominal ultrasound trauma recognition dataset, TRUST achieves a 9.63\% improvement over state-of-the-art methods with superior computational efficiency.

\section{Methodology}
\begin{figure}[t]
\includegraphics[width=\textwidth]{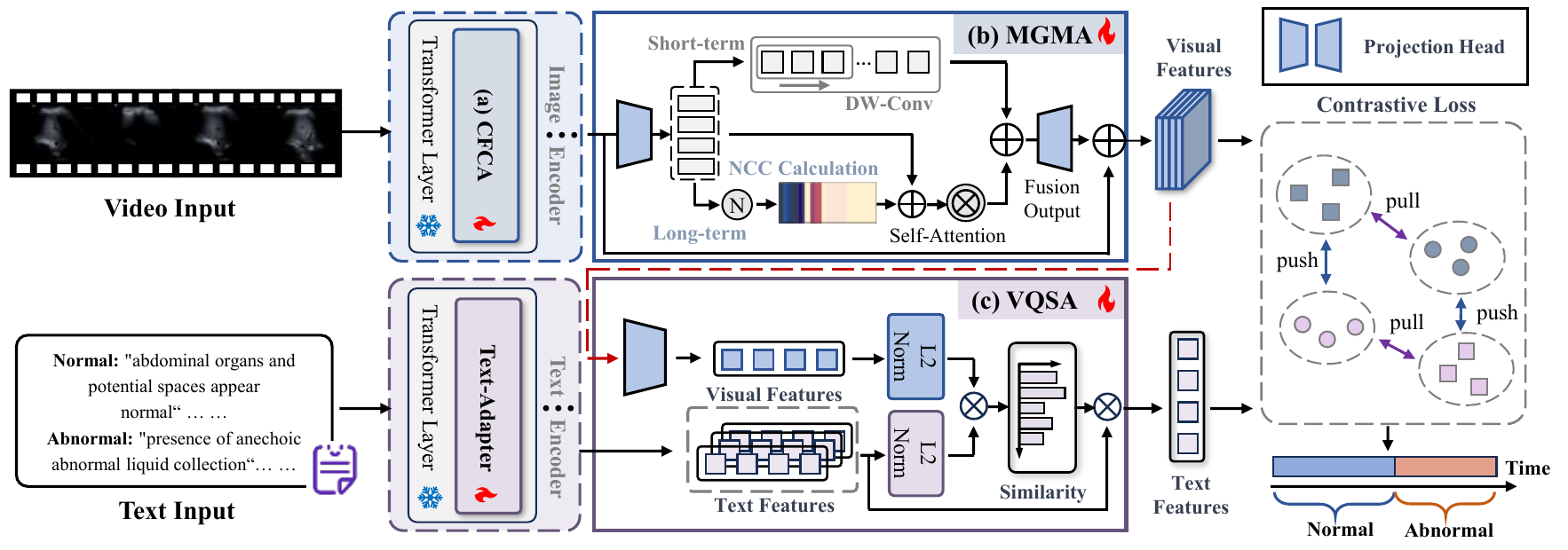}
\caption{Overview of the proposed TRUST. It includes: (a) CFCA, which extracts fine-grained spatial features; (b) MGMA, which models temporal dynamics in ultrasound videos; and (c) VQSA, which adaptively aggregates diverse textual descriptions.} \label{framework}
\end{figure}

As illustrated in Fig. \ref{framework}, TRUST adopts a CLIP-style contrastive learning paradigm designed to progressively refine feature representations across spatial, temporal, and semantic dimensions. Specifically, within the visual branch, the CFCA mechanism is systematically integrated into each transformer layer to facilitate more discriminative and fine-grained feature extraction. In addition, the MGMA module is incorporated to enhance temporal dependency modeling and improve the representation of dynamic patterns. On the other hand, in the textual branch, the Text-Adapter and VQSA strategies are integrated to strengthen semantic representation and promote more precise cross-modal alignment. 

\subsection{Cross-Frequency Collaborative Adapter (CFCA)}
Existing frequency-based approaches~\cite{zhang2024domesticating,liu2025dualadapter} typically decompose features into high- and low-frequency components for independent processing. Although this strategy enhances frequency-specific modeling, it may amplify sensitivity to speckle noise inherent in ultrasound images, thereby limiting its effectiveness in abdominal ultrasound analysis. To this end, we propose CFCA, which explicitly models mutual constraints between low- and high-frequency components to promote cross-frequency information exchange, thereby extracting more discriminative features under challenging ultrasound scenarios~\cite{ye2025escnet}.
\begin{figure}[t]
\centering
\includegraphics[width=.9\textwidth]{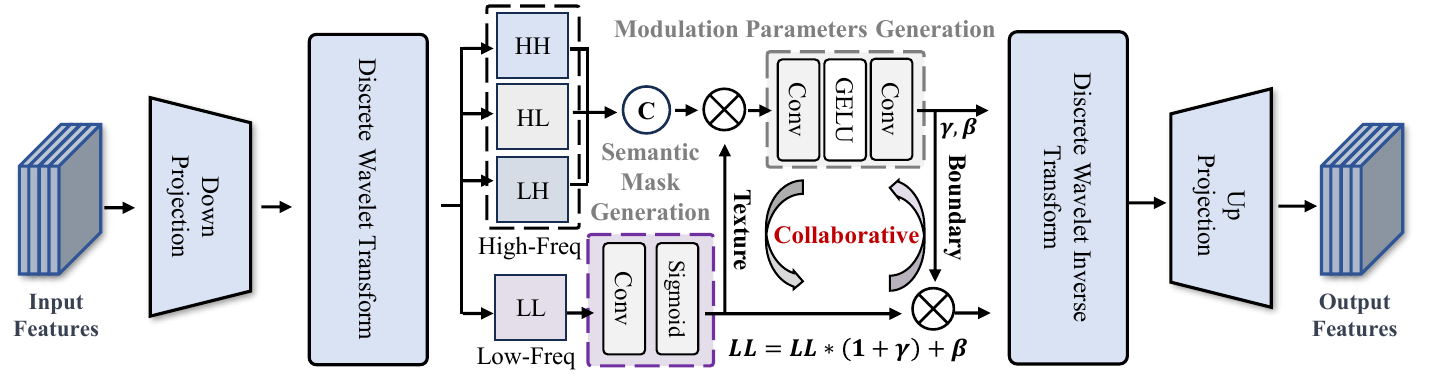}
\caption{Diagram of the CFCA module.} \label{CFCA_module}
\end{figure}

As illustrated in Fig.~\ref{CFCA_module}, the input features are first decomposed via wavelet transform into a low-frequency sub-band (LL) encoding coarse structural patterns, and three high-frequency sub-bands (HL, LH, HH) capturing fine-grained detail variations. Since the low-frequency sub-band carries global semantic certainty, we use it to generate a spatial mask that gates the high-frequency components, suppressing noise-induced spurious responses while preserving diagnostically relevant details:
\begin{equation}
    F_{High} = \sigma(Conv2D(F_{LL})) \odot Cat(F_{HL},F_{LH},F_{HH}),
\end{equation}
Conversely, high-frequency features provide spatially precise cues that refine the low-frequency component's representational scope, preventing semantic over-generalization.  We thus derive adaptive scaling $\gamma$ and offset $\beta$ from them to sharpen low-frequency discrimination:
\begin{equation}
    F_{Low} = F_{LL} \odot (1+\gamma)+\beta,
\end{equation}
Finally, the processed high-frequency and low-frequency features undergo inverse wavelet transformation and dimensionality projection to produce the final features $x_t=Proj(IWT(F_{Low}, F_{High}))$.
\subsection{Multi-Granularity Motion-Aware (MGMA) Module}
Ultrasound scanning exhibits dual temporal dynamics: smooth intra-view tissue motion and abrupt inter-view anatomical discontinuities from probe repositioning. Standard fixed-grid temporal layers~\cite{lea2017TCN,lea2016TCN2} apply uniform aggregation that inevitably conflates these two regimes, introducing noise at view boundaries. MGMA (Fig.~\ref{framework}(b)) disentangles them with a dual-branch architecture that enforces local continuity and calibrates global context via motion priors.

\textbf{Short-range Branch (Intra-view Consistency).} A depth-wise temporal convolution operates on the input sequence  $\mathbf{X}$ as a temporal low-pass filter, yielding $\mathbf{X}_S = {DWConv1D}(\mathbf{X})$, which suppresses frame-level jitter while preserving subtle pathological motion continuity within stable views.

\textbf{Long-range Branch (Inter-view Scanning Logic).} To capture global context while mitigating the disruption caused by anatomical discontinuities, we introduce Motion-Prior Position Encoding (MPE). The core insight is that semantic displacement serves as a robust surrogate for physical probe movement. We define the relative physical coordinate $p_t$ as the cumulative sum of semantic distances between adjacent frames within the causal sampling sequence. This mechanism dynamically stretches the temporal receptive field during rapid probe shifts and compresses it during stable scanning, aligning attention with the actual scanning logic:
\begin{equation}
p_t = \sum_{\tau=2}^{t} \left( 1 - \frac{\mathbf{x}_{\tau} \cdot \mathbf{x}_{\tau-1}}{| \mathbf{x}_{\tau} | | \mathbf{x}_{\tau-1} |} \right), \quad \mathbf{X}_{L} = \text{MSA} \left( \left\{ \mathbf{x}_{t} + \Phi(p_t) \right\}_{t=1}^T \right),
\end{equation}
Finally, the output integrates both granularities via a residual projection $\mathbf{X}_{Out} = (\mathbf{X}_S + \mathbf{X}_L)\mathbf{W}_{up} + \mathbf{X}$.

\subsection{Visual Query Semantic Aggregation (VQSA) Module}
Given the visual embedding from MGMA, the final step is cross-modal alignment. However, static text templates~\cite{wasim2023vita,wang2025vlpa,wang2021actionclip} fail to capture the significant intra-class variations caused by diverse scanning angles and probe pressure. Since different scanning conditions produce visually distinct manifestations of the same pathology, the optimal textual representation should be conditioned on each instance’s visual appearance. Based on this insight, VQSA uses the visual embedding as an \textit{anchor} to dynamically aggregate the most relevant descriptions from a diverse textual repository (Fig.~\ref{framework}(c)). Formally, let $\mathbf{v} = \text{Pool}(\mathbf{X}_{Out}) \in \mathbb{R}^d$ denote the final visual embedding. For the $c$-th category, we construct a fixed text bank using a structured slot-based prompting strategy, where each prompt is organized by category, modality, region, attribute, and temporal context. The generated candidates were manually verified according to FAST principles to exclude clinically ambiguous, redundant, or irrelevant descriptions before training. The resulting feature set $\mathcal{T}_c = \{ \mathbf{t}_c^k \}_{k=1}^K$ is extracted via the text encoder equipped with a standard bottleneck adapter~\cite{wang2024m2clip} for domain refinement. We then employ cross-modal attention where $\mathbf{v}$ serves as the Query and $\mathbf{t}_c^k$ as Keys and Values, dynamically synthesizing an instance-specific category prototype:
\begin{equation}
    \mathbf{p}_c(\mathbf{v}) = \sum_{k=1}^{K} {\text{Softmax} \left( \frac{\mathbf{v}^\top \mathbf{t}_c^k}{\sqrt{d}} \right)} \cdot \mathbf{t}_c^k,
\end{equation}
Finally, following the CLIP paradigm, the classification logits are predicted by computing the cosine similarity between the visual embedding $\mathbf{v}$ and the synthesized dynamic prototype $\mathbf{p}_c(\mathbf{v})$.

\subsection{Overall Loss}
To enhance discriminative capabilities for downstream tasks while preserving pre-trained knowledge, we designed a hybrid objective function:
\begin{equation}
    \mathcal{L}_{total} = \lambda \mathcal{L}_{con} + (1-\lambda) \mathcal{L}_{cls}.
\end{equation}
where $\lambda$ is a balancing coefficient, which is set to 0.5, $\mathcal{L}_{con}$ adopts the standard symmetric InfoNCE formulation to achieve cross-modal alignment, whilst $\mathcal{L}_{cls}$ employs a standard cross-entropy loss based on image-text similarity logits, explicitly optimizing classification accuracy.
\section{Experiments and Results}
\begin{table*}[t]
\caption{Comparison of results within the abdominal ultrasound video dataset. Frames denotes the number of sampled frames $\times$ sampling interval. Modal indicates the input modality for the model: image alone is denoted as $I$, while simultaneous image and text input is denoted as $I+T$.}
\label{tab1}
\centering
\fontsize{8}{10}\selectfont
\begin{tabular}{l|l|c|ccccc}
\hline
 Frames &Method & Modal & Accuracy & Precision & Recall & F1-score & Jaccard \\
\hline
\multirow{7}{*}{$4\times2$}&ST-Adapter{\scriptsize \textcolor{gray}{[NIPS'22]}} &I & 65.00 & 64.48&67.44 &63.33 &46.87\\
&AIM{\scriptsize \textcolor{gray}{[ICLR'23]}}  &I & 68.77&61.80 &60.57 &60.98 &46.31\\ 
&Surg-PEFT{\scriptsize \textcolor{gray}{[TMI'25]}} &I& \underline{70.76}&35.34 &50.00 &41.41 &35.34\\
&Vita-CLIP{\scriptsize \textcolor{gray}{[CVPR'23]}} &I+T&66.63 &54.20 &52.60  &51.68 &39.42\\
&M2-CLIP{\scriptsize \textcolor{gray}{[AAAI'24]}} &I+T&{69.69} &\underline{69.09} &\underline{72.99}&\underline{68.23} &\underline{52.24}\\
&VLPA-CLIP{\scriptsize \textcolor{gray}{[PR'25]}}& I+T&66.84&68.74&72.28&65.24&48.62\\ 
&TRUST{\scriptsize \textcolor{gray}{[Ours]}}& I+T &\textbf{81.36} &\textbf{76.89} &\textbf{81.00} &\textbf{77.89} &\textbf{64.32}\\ \hline

\multirow{7}{*}{$8\times2$}&ST-Adapter{\scriptsize \textcolor{gray}{[NIPS'22]}} &I & 66.74 & 59.28&58.84 &59.02 &44.18\\
&AIM{\scriptsize \textcolor{gray}{[ICLR'23]}}  &I & 70.76&35.19 &50.00 &41.31 &35.19\\ 
&Surg-PEFT{\scriptsize \textcolor{gray}{[TMI'25]}} &I& 65.26&56.22 &55.15 &55.23 &41.29\\
&Vita-CLIP{\scriptsize \textcolor{gray}{[CVPR'23]}} &I+T&71.06 &63.44 &51.80  &46.30 &37.89\\
&M2-CLIP{\scriptsize \textcolor{gray}{[AAAI'24]}} &I+T&65.56 &68.41 &71.92&65.67 &49.14\\
&VLPA-CLIP{\scriptsize \textcolor{gray}{[PR'25]}}& I+T&\underline{73.71}&\underline{71.68}&\underline{75.71}&\underline{71.73}&\underline{56.42}\\ 
&TRUST{\scriptsize \textcolor{gray}{[Ours]}}& I+T &\textbf{83.34} &\textbf{78.63} &\textbf{82.71} &\textbf{79.78} &\textbf{66.85}\\ \hline
\end{tabular}
\label{comparison}
\end{table*}
\subsection{Datasets and Experimental Settings} For the abdominal ultrasound video trauma detection task, we utilized a proprietary dataset consisting of 294 videos. All frames in these videos were resized to 224×224 resolution and annotated with labels indicating normal or abnormal stages. The videos were divided into training and test sets by patient, comprising 231 and 63 videos, respectively. For the number of texts corresponding to each category $K$, we set this value to $8$ in the reported experimental results. For the evaluation metrics, we first selected the Jaccard coefficient (tIoU), which directly relates to interval localization. We then chose metrics for fine-grained recognition, including Accuracy, Recall, Precision, and the F1 score. For all experiments, both the image and text encoders used the weights of CLIP ViT-B/16, with the pre-trained encoders kept frozen during training. All experiments were conducted on 4 RTX 4090 GPUs, using a batch size of 32 and training for 36 epochs. We employ the AdamW optimizer and use a base learning rate of 1e-3.

\subsection{Comparison with State-of-the-Art Methods.}
We compared our approach with two categories of state-of-the-art PEIVTL methods: unimodal methods, including ST-Adapter~\cite{pan2022st-adapter}, AIM~\cite{yang2023aim}, and SurgPEFT~\cite{yang2025surgpetl}; and multimodal methods, including Vita-CLIP~\cite{wasim2023vita}, M2-CLIP~\cite{wang2024m2clip}, and VLPA-CLIP~\cite{wang2025vlpa}. All methods were evaluated under identical conditions using two video sampling strategies: $4\times2$ and $8\times2$ (frames$\times$interval). As shown in Table \ref{comparison}, under the $4\times2$ setting, TRUST achieved an accuracy of 81.36\% and a Jaccard coefficient of 64.32\%, surpassing the second-best method by 10.6 and 12.08 percentage points, respectively. This significant performance advantage underscores the necessity of our motion-aware design. Traditional methods suffer from feature confusion due to inherent anatomical variations in ultrasound, whereas TRUST effectively leverages motion prior information to model both short- and long-range temporal information.
\begin{figure}[t]
    \centering
    \includegraphics[width=.9\linewidth]{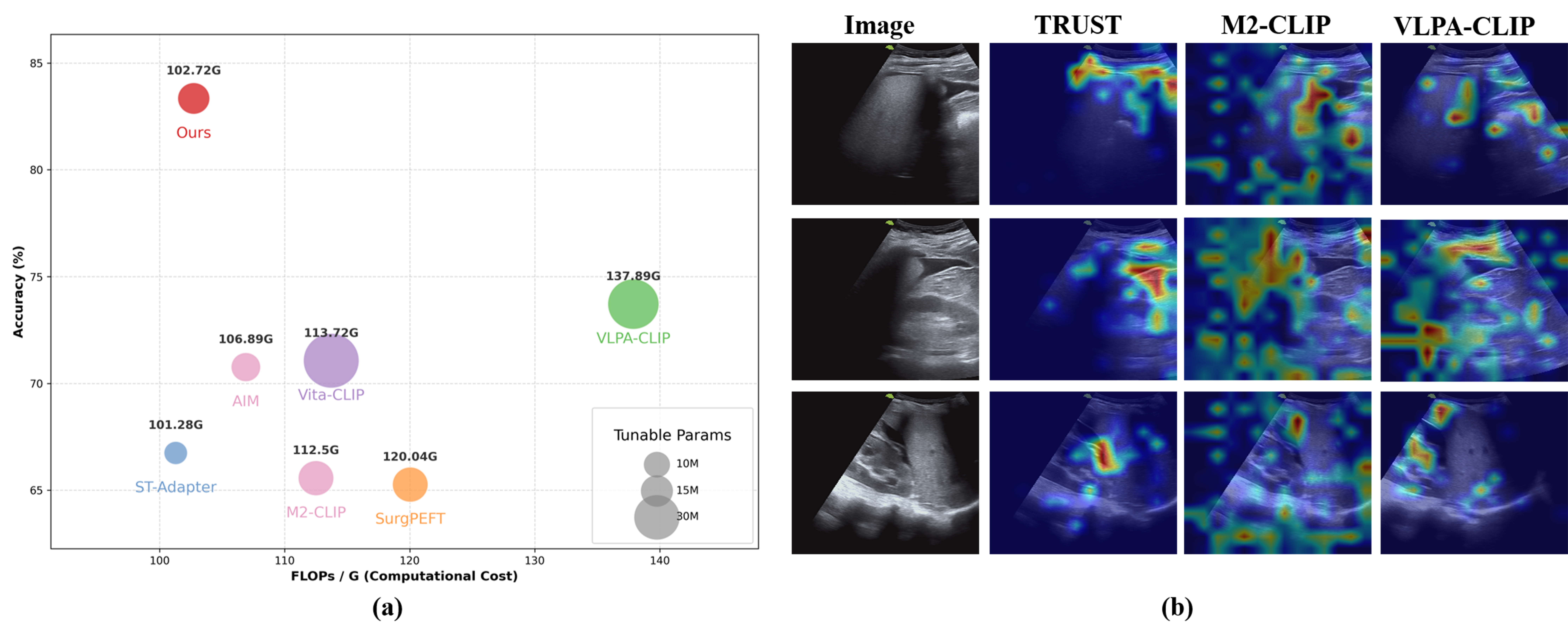}
    \caption{(a) Comparison of computational complexity under an 8×2 sampling strategy, where the circle size represents the model's trainable parameter scale; (b) Heatmap of text-image similarity generated using the CLIP model.}
    \label{vis_analysis}
\end{figure}

Moreover, Fig. \ref{vis_analysis} (a) illustrates the complexity comparison under the $8\times2$ strategy, where TRUST achieves optimal classification performance while maintaining highly competitive computational efficiency. Fig. \ref{vis_analysis} (b) presents a qualitative analysis. The heatmap, based on image-text similarity, demonstrates that TRUST captures fine-grained spatial information, enabling precise localization of traumatic lesions. In contrast, the comparison method exhibits noticeable attention divergence or erroneous activation toward background noise, further validating the effectiveness of TRUST’s cross-modal alignment.
\subsection{Ablation Study.}
\begin{table*}[t]
\caption{Experimental results of the ablation of each component of the model.}\label{tab1}
\centering
\fontsize{8}{10}\selectfont
\begin{tabular}{cccc|ccccc}
\hline
CFCA & Text-Adapter & MGMA & VQSA &Accuracy & Precision&Recall & F1-score&Jaccard\\
\hline
\ding{53} &\ding{53} &\ding{53} & \ding{53} & 67.50& 51.73 &50.62 &47.17  &37.27  \\
\checkmark &\ding{53} &\ding{53} & \ding{53} &70.96 &69.17  & 73.01& 68.66 & 52.78 \\
\checkmark &\checkmark &\ding{53} & \ding{53} &75.38 & 71.77 & 72.93& 72.27 & 57.68 \\
\checkmark &\checkmark &\checkmark & \ding{53} &76.36 & 73.05 &77.22 & 73.50& 58.64 \\ \hline
\checkmark &\checkmark &\checkmark & \checkmark &\textbf{83.34} &\textbf{78.63} &\textbf{82.71} &\textbf{79.78} &\textbf{66.85} \\
\hline
\end{tabular}
\label{ablation_study}
\end{table*}
Table \ref{ablation_study} presents the results of ablation experiments for each component. The baseline model achieved an accuracy of only 67.50\%. Incorporating CFCA increased accuracy to 70.96\%, highlighting the crucial role of enhancing fine-grained spatial features. Adding Text-Adapter and MGMA on this foundation further improved model performance to 76.36\%, demonstrating the effectiveness of text adaptation and motion perception-based temporal modeling. Most notably, the introduction of VQSA resulted in a significant performance boost, raising the accuracy and Jaccard score to 83.34\% and 66.85\%, respectively.

\begin{figure}[t]
    \centering
    \includegraphics[width=\linewidth]{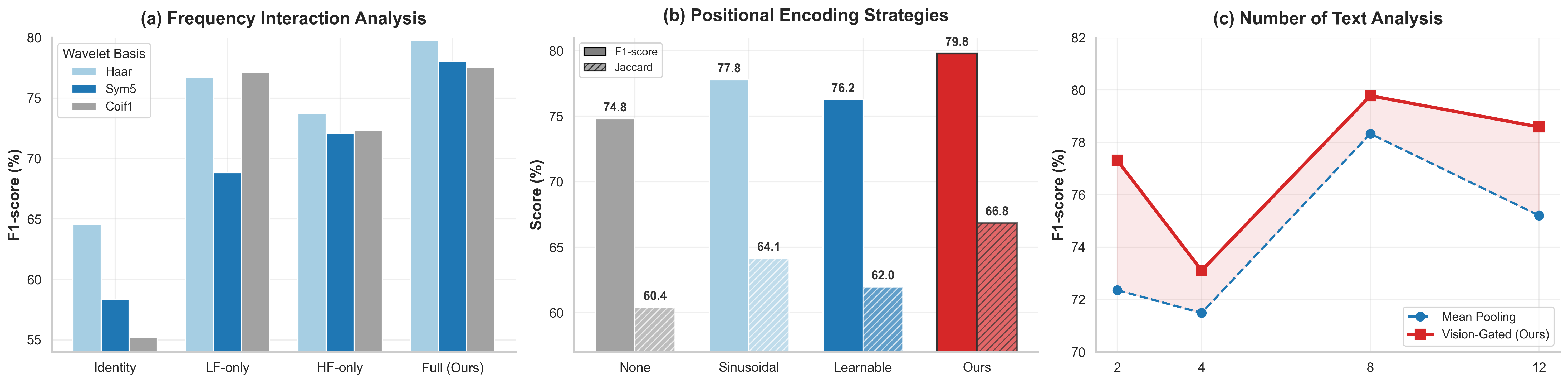}
    \caption{Three analytical experiments conducted using an 8×2 sampling strategy: (a) analysis of frequency-domain interaction patterns and wavelet basis functions; (b) evaluation of the impact of different positional encoding schemes; and (c) investigation of the influence of varying text quantities and fusion strategies.}
    \label{ablation_fig}
\end{figure}

Furthermore, we conducted more detailed analytical experiments on each proposed module, with the results presented in Fig. \ref{ablation_fig}. For CFCA, robustness was demonstrated across different wavelet basis functions, whereas interaction using sub-bands alone yielded suboptimal performance. Regarding positional encoding, our proposed motion-aware encoding achieved the best results. In terms of VQSA, our multi-text fusion strategy significantly outperformed mean pooling.

\section{Conclusion}
In this paper, we propose TRUST, a parameter-efficient image-to-video transfer learning framework specifically designed for abdominal ultrasound trauma detection. By integrating fine-grained spatial feature extraction, multi-granularity temporal information aggregation, and enhanced cross-modal alignment, our approach achieves highly efficient and accurate trauma detection. Notably, the method introduces a novel motion-aware positional encoding that effectively addresses semantic discontinuities caused by significant anatomical variations during ultrasound scanning, providing a new perspective for ultrasound video analysis tasks.
\subsubsection{\ackname} This work was supported in part by the National Natural Science Foundation of China (62371121), and in part by the Jiangsu Provincial Key Research and Development Program, China (BE2022827).
\subsubsection{\discintname}
The authors have no competing interests to declare that are relevant to the content of this article.

%
%
%
%

\bibliographystyle{splncs04} 
\bibliography{refs}          






\end{document}